\documentclass[journal]{IEEEtran}
\usepackage{amssymb,amsfonts}
\usepackage[cmex10]{amsmath}
\usepackage{graphicx}
\usepackage{algorithm}
\usepackage{algorithmic}
\usepackage{svg}
\usepackage{graphicx}
\usepackage{booktabs}
\usepackage{multirow}
\usepackage{url}
\ifCLASSINFOpdf
\else
\fi

\begin{document}

\title{AdaptStress: Online Adaptive Learning for Interpretable and Personalized Stress Prediction Using Multivariate and Sparse Physiological Signals}

\author{Xueyi Wang,~\IEEEmembership{}
        Claudine J. C. Lamoth,~\IEEEmembership{}
        Elisabeth Wilhelm,~\IEEEmembership{}
\thanks{Xueyi Wang is with the Engineering and Technology institute Groningen (ENTEG), University of Groningen, Groningen, 9742AG, the Netherlands. e-mail: (xueyi.wang@rug.nl).}
\thanks{Claudine J. C. Lamoth are with Department of Human Movement Sciences, University Medical Center Groningen, Groningen, Netherlands. e-mail: (c.j.c.lamoth@umcg.nl)}
\thanks{Elisabeth Wilhelm are with the Engineering and Technology institute Groningen (ENTEG), University of Groningen, Groningen, 9742AG, the Netherlands. e-mail: (e.wilhelm@rug.nl)}
\thanks{}}

\markboth{}%
{Shell \MakeLowercase{\textit{et al.}}: Bare Demo of IEEEtran.cls for IEEE Journals}

\maketitle

\begin{abstract}
Continuous stress forecasting could potentially contribute to lifestyle interventions. This paper presents a novel, explainable, and individualized approach for stress prediction using physiological data from consumer-grade smartwatches. We develop a time series forecasting model that leverages multivariate features, including heart rate variability, activity patterns, and sleep metrics, to predict stress levels across 16 temporal horizons (History window: 3, 5, 7, 9 days; forecasting window: 1, 3, 5, 7 days). Our evaluation involves 16 participants monitored for 10-15 weeks. We evaluate our approach across 16 participants, comparing against state-of-the-art time series models (Informer, TimesNet, PatchTST) and traditional baselines (CNN, LSTM, CNN-LSTM) across multiple temporal horizons. Our model achieved performance with an MSE of 0.053, MAE of 0.190, and RMSE of 0.226 in optimal settings (5-day input, 1-day prediction). A comparison with the baseline models shows that our model outperforms TimesNet, PatchTST, CNN-LSTM, LSTM, and CNN under all conditions, representing improvements of 36.9\%, 25.5\%, and 21.5\% over the best baseline. According to the explanability analysis, sleep metrics are the most dominant and consistent stress predictors (importance: 1.1, consistency: 0.9-1.0), while activity features exhibit high inter-participant variability (0.1-0.2). Most notably, the model captures individual-specific patterns where identical features can have opposing effects across users, validating its personalization capabilities. These findings establish that consumer wearables, combined with adaptive and interpretable deep learning, can deliver relevant stress assessment adapted to individual physiological responses, providing a foundation for scalable, continuous, explainable mental health monitoring in real-world settings.
\end{abstract}


\IEEEpeerreviewmaketitle

\section{Introduction}


Chronic stress not only triggers the onset of depression and anxiety but also worsens their progression by affecting critical brain regions responsible for mood regulation \cite{mcewen2007physiology}. While current technologies can effectively monitor stress levels, the field lacks robust predictive models capable of forecasting individual stress patterns, leaving a critical gap in preventive mental healthcare.



The widespread adoption of wearable devices has revolutionized stress monitoring by continuously capturing physiological signals such as heart rate variability and electrodermal activity \cite{cohen2007psychological}. However, a fundamental limitation persists—while algorithms for estimating current stress from wearable data are well-established, predictive models for individual stress forecasting remain scarce. This gap is particularly problematic for long-term therapies and preventive strategies, where anticipating stress patterns could enable timely interventions before critical episodes occur. Current stress-monitoring efforts face three critical obstacles. First, comprehensive data collection remains burdensome, often yielding sparse and incomplete datasets that fail to capture the full spectrum of stress-related factors \cite{taylor2017personalized}. Secondly, the relationships among stress-related variables exhibit nonlinear patterns and complex interactions that confound standard modeling approaches, particularly when dealing with sparse, heterogeneous data \cite{oken2015systems, sano2018identifying}. Finally, it is essential for clinical deployment, as therapists need to understand and trust the system's predictions to effectively integrate them into intervention strategies \cite{rabbi2015mybehavior, ribeiro2016should}. Stress forecasting plays a vital role in enabling interventions such as stress management training and workplace adjustments, which aim to minimize the adverse impacts of stress \cite{haque2024state}. 


This work presents a novel explainable and personalized stress forecasting framework that predicts individual stress trajectories days in advance from data acquired with wearable sensors, enabling proactive intervention while maintaining clinical interpretability. Our approach introduces three key innovations: (1) a multivariate temporal sparse framework that learns compact, interpretable representations of stress-related variables while handling missing data and capturing temporal dynamics and cross-variable interactions; (2) structured priors and domain-specific constraints that embed established stress physiology knowledge to ensure clinically meaningful outputs; and (3) adaptive feature selection and importance ranking techniques that identify the most relevant variables for each individual, reducing data collection burden while maintaining predictive power.

This paper is organized as follows. Section II reviews related work in stress monitoring and wearable computing. We introduced data description, participant recruitment, data collection protocols, and processing in Section III and present our multivariate temporal framework and its theoretical foundations in Section IV. Section V reports results from our experiments, comparing our approach against baseline methods and analyzing feature importance patterns. Finally, Section VI concludes with future research directions for explainable stress forecasting systems.

\vspace{0.2in}
\begin{figure*}
    \centering
    \includegraphics[width=1\textwidth]{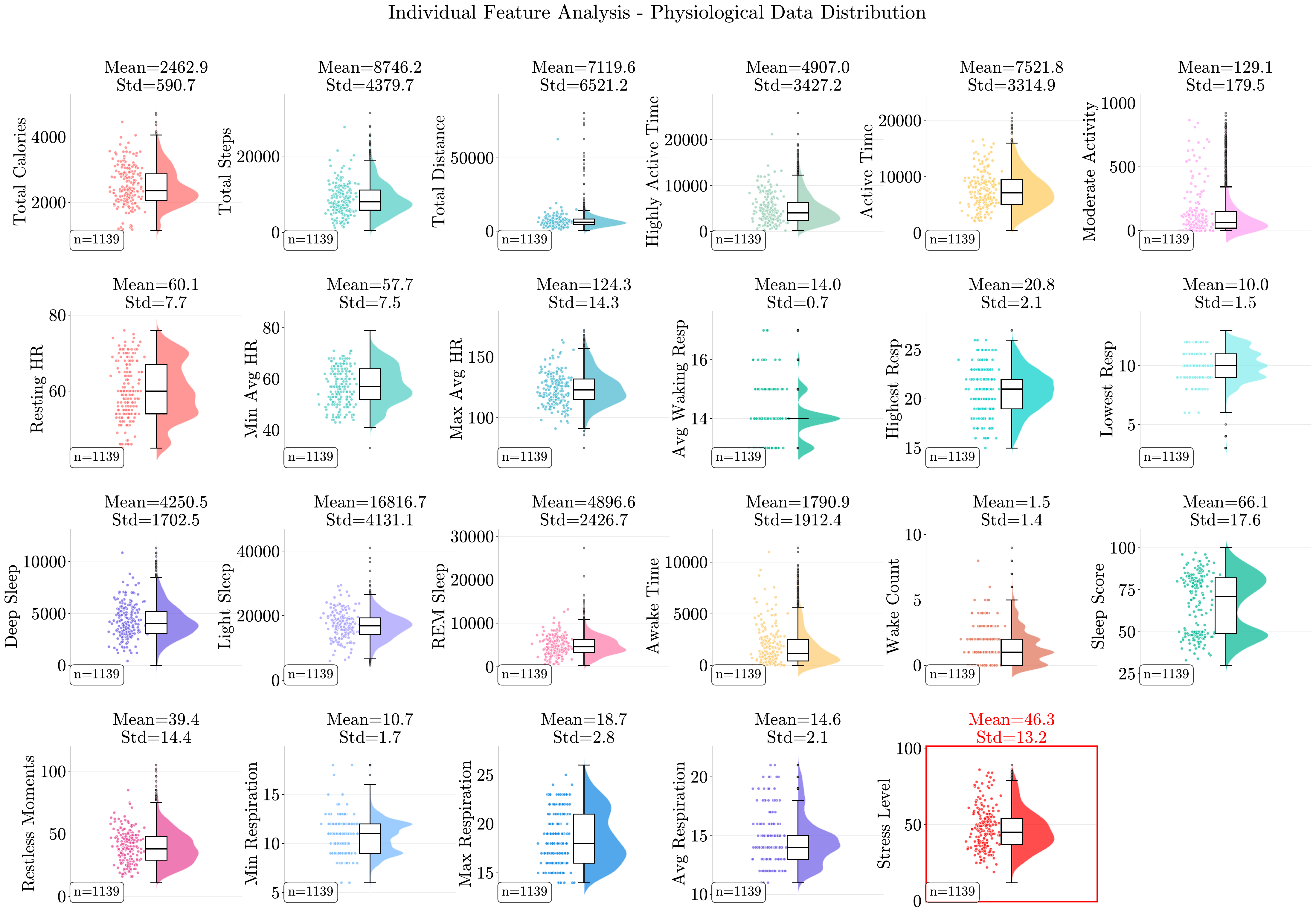}
    \caption{Individual feature analysis of physiological data distribution.}
    \label{fig:box-plot}
\end{figure*}

\section{Related Work}
Stress can be measured through various methods, each capturing different aspects of the stress response. The state-of-the-art in stress measurement includes both subjective and objective approaches, as well as the use of advanced technologies for continuous and real-time monitoring. Self-report questionnaires are widely used to assess perceived stress levels and the emotional, cognitive, and behavioral aspects of stress. The Perceived Stress Scale (PSS) \cite{cohen1983global} and the Depression Anxiety Stress Scales (DASS) \cite{lovibond1995structure} are commonly employed instruments that measure stress based on an individual's subjective evaluation of their experiences. 


Wearable sensors, particularly wrist-worn devices, enable continuous monitoring of physiological stress indicators such as skin conductance, temperature, and heart rate \cite{gjoreski2017continuous}. Smartphone-based methods leverage built-in sensors and user interactions to infer stress levels based on patterns of phone usage, physical activity, and sleep \cite{sano2018identifying}. Researchers~\cite{garcia2015automatic} collected smartphone accelerometer data from approximately 30 subjects over eight weeks in real working environments to detect stress-related behavior patterns, with participants self-reporting perceived stress levels three times daily. Statistical classification models achieved 71\% accuracy for user-specific stress prediction and 60\% accuracy for similar-users models using only accelerometer sensor data. These approaches allow for unobtrusive and real-time stress monitoring in natural environments.

Machine learning techniques have been increasingly applied to multimodal datasets to enhance the accuracy and robustness of stress predictions \cite{garcia2015automatic, giannakakis2019review}. Support Vector Machines (SVMs) were successfully used with feature selection to predict non-remitting PTSD in 957 trauma survivors assessed within 10 days post-trauma and followed for 15 months. The ML feature selection identified 16 key predictors that achieved AUC = 0.77 for forecasting non-remitting PTSD (comparable to using all features, AUC = 0.78)~\cite{galatzer2014quantitative}, substantially outperforming prediction based solely on Acute Stress Disorder symptoms (AUC = 0.60). However, these models often lack explainability, which is crucial for fostering user trust and adherence in lifestyle interventions. Explainable AI (XAI) methods have been proposed to provide interpretable explanations for machine learning predictions \cite{adadi2018peeking, arrieta2020explainable}. A multimodal stress and affective states prediction model was developed using wrist-worn (PPG, GSR, ACC, TEMP) and chest-worn (ECG, EMG, GSR, RESP, TEMP) sensors from the WESAD dataset with 15 participants. They compared six conventional machine learning classifiers (SVM, LDA, KNN, NN, Decision Tree, Random Forest) for 4-class classification (baseline, stress, amusement, meditation) using leave-one-out cross-validation. Random Forest achieved the best performance with the accuracy of 95.54±2.14\% for wrist-worn sensors and 97.15±2.67\% for chest-worn sensors, with chest-worn ECG and GSR being the most discriminative modalities~\cite{gupta2023multimodal}. Despite the common goal of these studies, the number of days required to achieve reliable predictions and the predictive time frames vary. Sequential Minimal Optimization for Regression (SMOreg) and Linear Regression to forecast stress levels in families of cancer patients using data from 217,067 patients (1998-2010). Linear Regression achieved better performance (RMSE = 1076.15 vs 1223.75 for SMOreg), predicting an average family stress level of 72.71\% for cancer patients' relatives~\cite{adeel2022stress}. In educational settings, six machine learning classification algorithms (Naive Bayes, Linear Regression, Multi-layer Perceptron, Bayes Net, J48, and Random Forest) were used to predict stress levels in over 200 university students assessed using the Perceived Stress Scale. Random Forest achieved the highest accuracy of 94.73\% for stress prediction among student populations~\cite{sharma2020stress}.


The issue of generalizability across individuals was tested by developing more reliable and ecologically valid stress measurement techniques, as well as personalized stress models that adapt to individual differences and situational factors~\cite{epel2018more}. Individual differences in stress reactivity and coping strategies necessitate personalized models that adapt to each user's unique characteristics \cite{epel2018more}. Personalized machine learning models were developed to predict tomorrow's mood, stress, and health levels, including a Multitask Learning Deep Neural Network (MTL-DNN) with person-specific layers and a Domain Adaptation Gaussian Process (DA-GP)~\cite{jaques2017predicting}. They used 30-day data from 69 participants with 343 features from wearable sensors, smartphones, and surveys, and treated each person's prediction as a separate task. Their personalized models achieved 13-22\% lower prediction error compared to generic models, demonstrating the importance of accounting for individual differences in affective state prediction. 



In this paper, we propose an explainable and personalized stress forecasting framework that leverages multivariate sparse data in lifestyle interventions. Our approach combines XAI methods and domain adaptation to provide interpretable and individualized stress predictions, effectively addressing the limitations of existing stress measurement and prediction techniques.

\section{Data Description and Processing}


It was conducted in accordance with the Declaration of Helsinki and Dutch regulations governing research involving human participants. The Central Ethics Review Board for non-WMO studies (CTc UMCG), which oversees investigations outside the scope of the Medical Research Involving Human Subjects Act (WMO), reviewed and approved the study (study register number 18021). Written informed consent was obtained from each participant prior to enrollment.
We recruited 17 volunteers (median age: 55.5 years, IQR: 16.25; median BMI: 28.22~kg/m$^2$ BMI (IQR: 10.74) from two community health programs in northern Netherlands: a supervised weekly walking group and a 10-15 week lifestyle intervention combining nutrition education, physical activity, and meditation. One participant withdrew after three weeks, yielding 16 complete participants. 

All participants wore Garmin Vivosmart 5 devices for 10-15 weeks, with daily removal for charging following manufacturer guidelines. Research staff provided technical support through weekly walking sessions and bi-weekly site visits while minimizing disruption to participants' routines.

All data are securely stored in the University of Groningen's protected offline Virtual Research Workspace (VRW). The protocol involves data preprocessing within VRW to remove sensitive information, followed by download and local analysis of preprocessed data. More information could be found in our previous work·~\cite{wang2024multivariate}.

\begin{table}[htbp]
      \centering
      \caption{Feature List with Abbreviations, ranges, and units}
      \begin{tabular}{p{3.5cm}p{0.7cm}p{1.6cm}p{1.3cm}}
      \hline
      \textbf{Feature Name} & \textbf{Abbrev.} & \textbf{Range} &
  \textbf{Unit} \\
      \hline
      Total Kilocalories & TK & 381-4726 & kcal\\
      Total Steps & TS & 46-31374 & steps \\
      Total Distance Meters & TD & 31-80528 & meters \\
      Highly Active Seconds & HA & 0-25768 & seconds \\
      Active Seconds & ACS & 0-21357 & seconds\\
      Moderate Intensity Minutes & MI & 0-961 & minutes\\
      Resting Heart Rate & RH & 45-76 & bpm \\
      Minimum Average Heart Rate & MIR & 33-81 & bpm \\
      Maximum Average Heart Rate & MXR & 75-172 & bpm \\
      Average Respiration Value & AWR & 13-17 & brpm\\
      Highest Respiration Value & HRV & 15-27 & brpm\\
      Lowest Respiration Value & LRV & 3-13 & brpm\\
      Deep Sleep Seconds & DS & 0-11340 & seconds\\
      Light Sleep Seconds & LS & 600-41100 & seconds\\
      REM Sleep Seconds & RS & 300-27420 & seconds\\
      Awake Sleep Seconds & AWS & 0-12360 & seconds\\
      Awake Count & AC & 0-9 & count\\
      Sleep Overall Score & SOS & 13-100 & arbitrary\\
      Restless Moment Count & RMC & 2-105 & count\\
      Lowest Respiration & LR & 6-18 & brpm\\
      Highest Respiration & HRS & 14-26 & brpm\\
      Average Respiration & AR & 11-21 & brpm \\
      \textbf{Stress Score} & \textbf{SS} & \textbf{0-95} &
  \textbf{arbitrary}\\
      \hline
      \end{tabular}
      \label{tab:features}
  \end{table}

\begin{figure}
    \centering
    \includegraphics[width=0.5\textwidth]{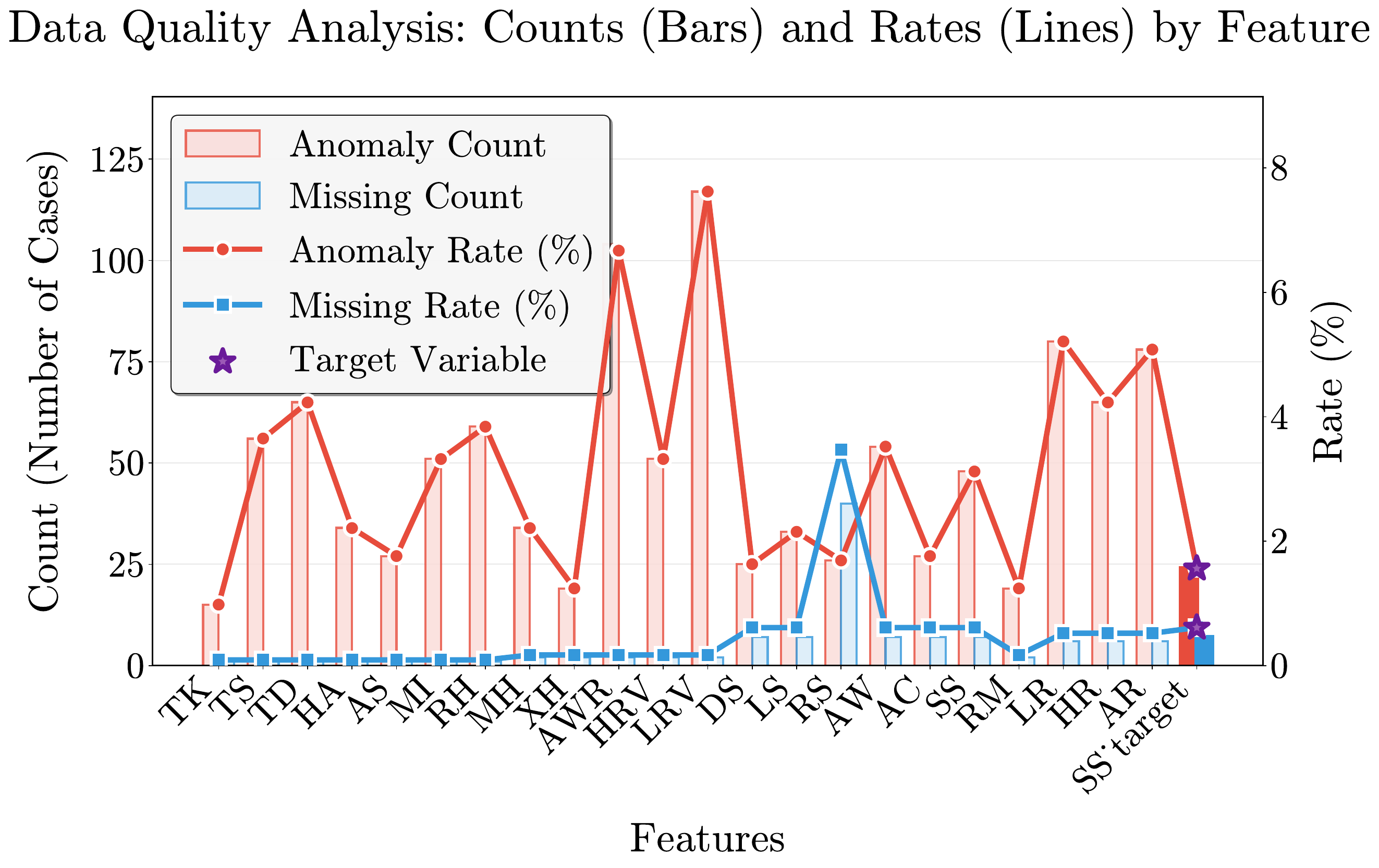}
    \caption{Missing and anomaly numbers and rates for all features}
    \label{fig:missing-anomaly}
\end{figure}

\subsection{Features}
A total number of 23 features can be downloaded from the manufacturer of the smartwatch: Total Kilocalories (TK), Total Steps (TS), Total Distance Meters (TD), Highly Active Seconds (HA), Active Seconds (ACS), Moderate Intensity Minutes (MI), Resting Heart Rate (RH), Minimum Average Heart Rate (MIR), Maximum Average Heart Rate (MXH), Average Walking Respiration Value (AWR), Highest Respiration Value (HRV), Lowest Respiration Value (LRV), Deep Sleep Seconds (DS), Light Sleep Seconds (LS), REM Sleep Seconds (RS), Awake Sleep Seconds (AWS), Awake Count (AC), Sleep Overall Score (SS), Restless Moment Count (RM), Lowest Respiration (LR), Highest Respiration (HRS), and Average Respiration (AR). All features used in this work were calculated by the Garmin Connect app. These features encompass multiple physiological domains, including physical activity metrics, respiratory measurements, and sleep quality indicators.

We implement a comprehensive feature selection framework to reduce our 22 physiological and activity features to 15 most relevant predictors, achieving 32\% dimensionality reduction while maintaining predictive performance. Our approach employs four methods: correlation-based selection using Pearson coefficients (threshold 0.05), Recursive Feature Elimination with Random Forest, mutual information selection for non-linear dependencies, and an ensemble voting mechanism combining all three approaches. 

\subsection{Analysis the all features}
The box plot of Figure~\ref{fig:box-plot} presents the distribution of 23 physiological and activity features collected and extracted from Garmin smartwatches used for stress prediction. The activity-related features demonstrate substantial variability: Total Steps (mean=8,746.2, std=4,379.7), Total Distance (mean=7,119.6m, std=6,521.2m), and Moderate activity (mean=129.1 and Std=179.5) exhibit right-skewed distributions with considerable inter-participant differences, reflecting diverse activity patterns in the cohort. Time-based activity metrics show similar heterogeneity, with Highly Active Time averaging 4,907.0 seconds (std=3,427.2) and Active Time averaging 7,521.8 seconds (std=3,314.9).

Heart rate features display more normalized distributions with lower coefficients of variation: Resting HR (mean=60.1 bpm, std=7.7), Min Avg HR (mean=57.7 bpm, std=7.5), and Max Avg HR (mean=124.3 bpm, std=14.3) show relatively tight clustering around their means, indicating consistency across participants. Respiratory metrics follow similar patterns, with Avg Waking Resp maintaining a narrow distribution (mean=14.0 breaths/min, std=0.7).

Sleep architecture features reveal interesting patterns: Light Sleep dominates total sleep time (mean=16,816.7s, std=4,131.1s), while Deep Sleep (mean=4,250.5s, std=1,702.5s) and REM Sleep (mean=4,896.6s, std=2,426.7s) show proportionally smaller durations. The Sleep Score (mean=66.1, std=17.6) demonstrates a near-normal distribution, suggesting balanced sleep quality across the cohort. Notably, the target variable Stress Level (mean=46.3, std=13.2) exhibits a relatively symmetric distribution, providing a well-balanced prediction target without severe class imbalance issues. The prediction target, stress score (0-100), is primarily derived from a combination of HR and HRV data and calculated using Firstbeat Analytic~\cite{kuula2021heart}. The violin plots reveal that several features, particularly activity-related ones, contain multiple modes, suggesting distinct behavioral patterns or participant subgroups that may benefit from personalized modeling approaches. 


\subsection{Missing Data Imputation}
Incomplete data can significantly distort analysis results, potentially leading to misleading interpretations \cite{wu2020personalized}. In the original dataset, missing data are represented by NaN, -1, and -2. Anomaly detection was performed using two complementary approaches tailored to the characteristics of different physiological features. For general physiological metrics (e.g., heart rate, respiration, sleep, and activity features), outliers were identified using the Interquartile Range (IQR) method, where data points falling below $Q_1 - 1 \times IQR$ or above $Q_3 + 1 \times IQR$ were flagged as anomalies and replaced with NaN values. For stress-related features, a rolling mean-based approach was employed, where values deviating beyond a predefined threshold from the local moving average, as well as zero values indicating sensor artifacts, were marked as anomalous. Figure~\ref{fig:missing-anomaly} reveals data quality challenges across our 23 features, with respiratory features showing the highest anomaly rates (LRV at 7.6\%, AWR at 6.7\%). Activity features demonstrate moderate anomaly rates (TD: 4.2\%, TS: 3.7\%), while heart rate features show anomaly rates below 3.8\%. RS shows the highest missing rate at 3.5\% and other sleep architecture features at ~0.6\%, indicating systematic data loss during specific monitoring sessions. The target variable (Stress) maintains good quality with only 1.6\% anomalies and 0.6\% missing values, ensuring robust labels for training and highlighting the need for feature-specific preprocessing strategies. To enhance the overall quality of the dataset, we performed imputation using the SimpleImputer from the scikit-learn library. This imputation process aims to provide a more complete and reliable dataset for subsequent analysis, while acknowledging the limitations imposed by features with extensive missing data. However, the "Hydration" feature was excluded from our analysis. This feature requires manual input by users, and very few participants consistently provided this information daily, resulting in an exceptionally high rate of missing values. Data from each participant's first and last days are removed due to insufficient wearing time.

\section{Methods}

\subsection{Domain shifts}
Traditional machine learning methods operate under the assumption that the training, validation, and testing data are from the same distribution~\cite{moreno2012unifying, coleman2021forecasting}. This assumption, however, is often unrealistic, particularly when dealing with heterogeneous data collected from various conditions and participants~\cite{chen2024test}. Individual variability creates significant challenges for cross-participant generalization. 

\begin{figure}[!t]
    \centering
    \includegraphics[width=0.48\textwidth]{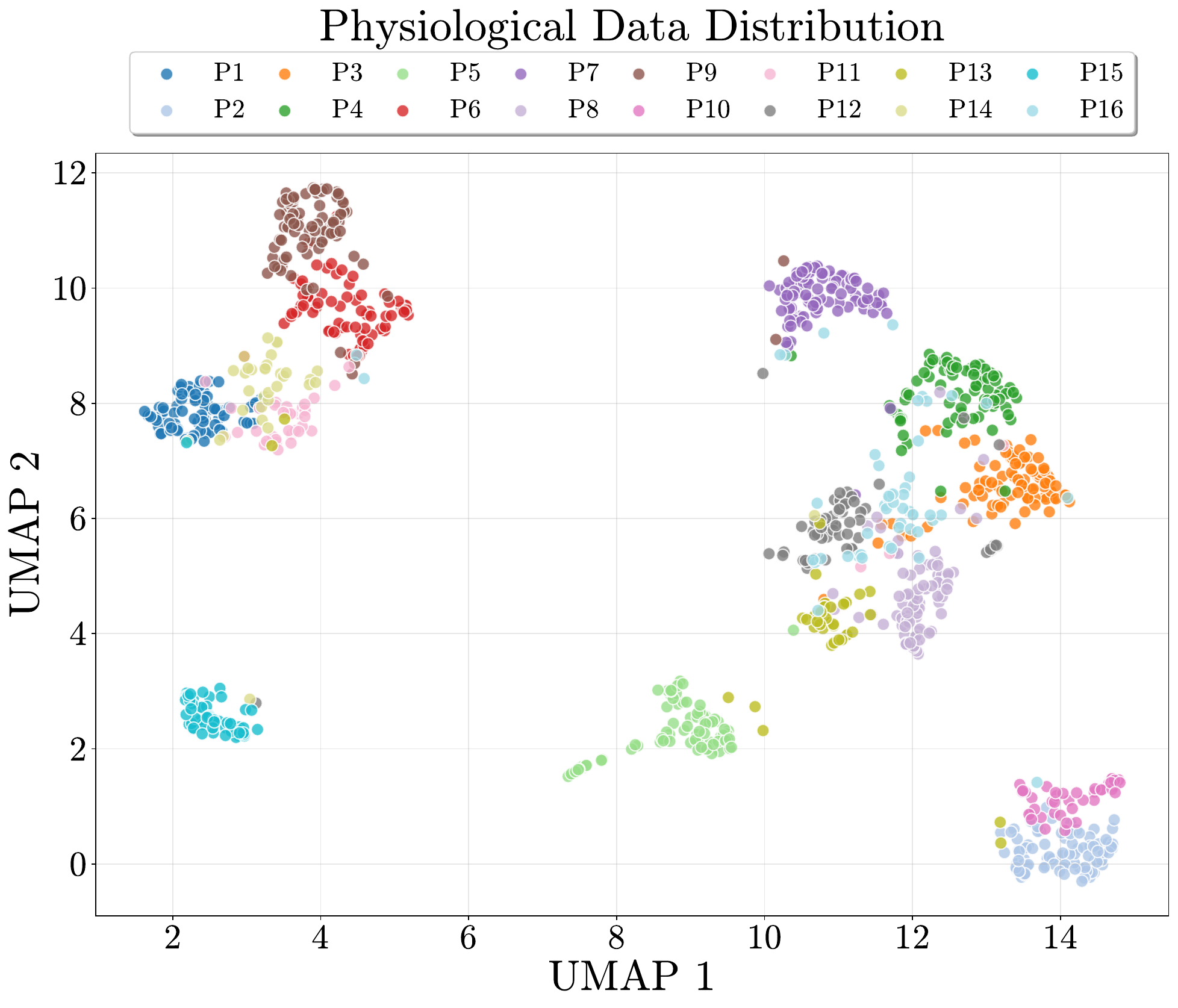}
    \caption{Domain shifts illustrated by UMAP analysis for features of each participant}
    \label{fig:umap}
\end{figure}

Figure~\ref{fig:umap} demonstrates pronounced domain shifts of our dataset across various participants through Uniform Manifold Approximation and Projection (UMAP) visualization, revealing distinct clustering patterns with clear spatial segregation. Each participant occupies unique manifold regions, confirming substantial inter-participant heterogeneity. Three primary clusters emerge: a lower-left cluster containing P15 with compact and consistent patterns; a central cluster dominated by P5; and a right-side cluster with P3, P14, and P16 forming separate sub-clusters. Participant clusters exhibit minimal overlap, with inter-participant transition zones of less than 0.5 UMAP units compared to intra-participant spreads exceeding 2-3 units, confirming the presence of substantial domain shifts.

\subsection{Models}
\begin{figure*}[t]
    \centering
    \includegraphics[width=1\textwidth]{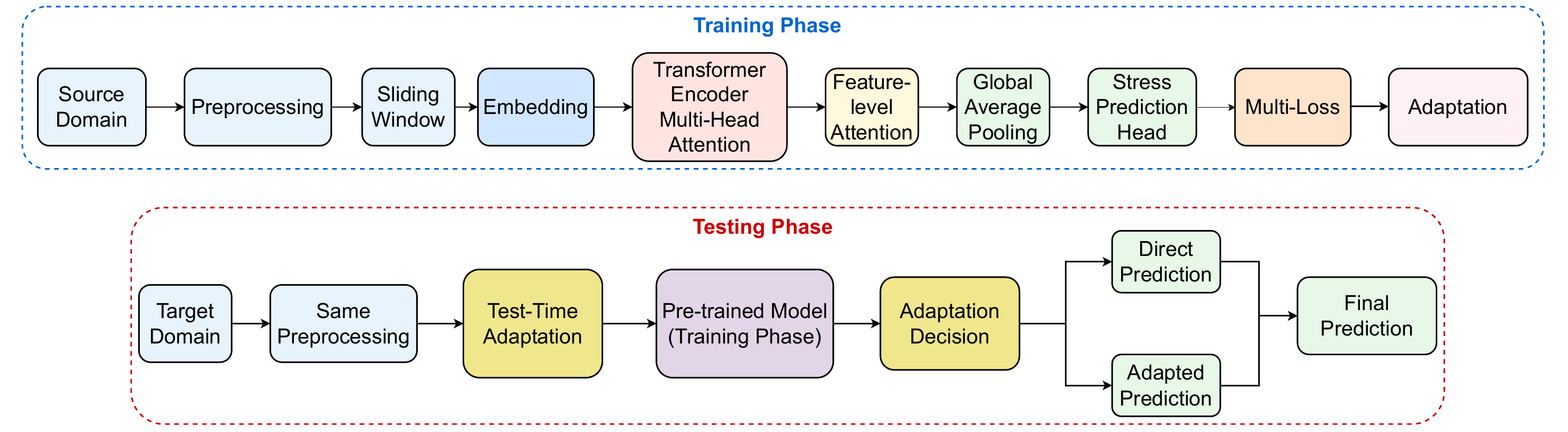}
    \caption{Diagram of AdaptStress model used in this work}
    \label{fig:flowchart}
\end{figure*}

Our proposed AdaptStress model is a flexible, interpretable, and domain-adaptive neural architecture for stress prediction, where test-time adaptation can be optionally enabled depending on deployment requirements across heterogeneous participant populations, as shown in Algorithm~\ref{alg:our-model} and Figure~\ref{fig:flowchart}. The model employs a multi-layer Transformer encoder, 8 attention heads, and 128-dimensional feature embeddings to capture complex temporal dependencies in 15-dimensional physiological time series, including heart rate variability, respiratory patterns, activity metrics, sleep quality indicators, and circadian rhythm features. The architecture integrates three key innovations: 1) a feature-level attention mechanism that dynamically weights the importance of different physiological signals based on their relevance to stress prediction, enabling adaptive feature selection across diverse participants; (2) a domain adaptation module incorporating adversarial training to minimize inter-participant domain shift during training on 14 source participants in the Leave-One-Out framework; and (3) a selective Test-Time Adaptation mechanism that computes domain similarity using statistical, distributional, and confidence-based metrics to determine whether adaptation is beneficial for each test participant, thereby avoiding performance degradation and computational overhead when the pre-trained model is already well-suited to the target domain. The architectural complexity is justified by the need to address the substantial domain distribution mismatches that arise in heterogeneous participant populations, as detailed in the following analysis.

\subsubsection{Domain Adaptation in Training}
Our AdaptStress model employs adversarial domain adaptation to address the fundamental challenge of inter-participant variability in stress prediction. Each individual represents a distinct domain with unique characteristics, violating the traditional assumption that source domain $\mathcal{D}_s = \{(\mathbf{x}_i^s, y_i^s)\}_{i=1}^{n_s}$ and target domain $\mathcal{D}_t = \{(\mathbf{x}_i^t)\}_{i=1}^{n_t}$ follow identical distributions, i.e., $P_s(\mathbf{x}) = P_t(\mathbf{x})$, where $\mathbf{x} \in \mathbb{R}^{T \times d}$ represents physiological time series and $y \in \mathbb{R}$ denotes stress levels. However, individual differences in stress reactivity, circadian rhythms, and baseline physiological patterns result in significant distribution shifts where $P_s(\mathbf{x}) \neq P_t(\mathbf{x})$. To mitigate this domain gap, our framework incorporates three key components: a feature extractor $G_f$ (Transformer encoder), a stress predictor $G_y$, and a domain discriminator $G_d$. The training objective combines stress prediction accuracy with adversarial domain confusion. This adversarial mechanism enables the model to learn domain-invariant yet task-discriminative representations, significantly improving cross-participant generalization capability.


\subsubsection{Selective Test-Time Adaptation}
Test-time adaptation (TTA) refines pre-trained models using incoming test data to address distributional shifts during inference~\cite{Liang2022}. Our selective Test-Time Adaptation (TTA) framework addresses the domain shift challenge during inference when encountering a new test participant with potentially different characteristics. Unlike traditional domain adaptation that requires access to both source and target data during training, TTA performs online model adaptation using only unlabeled target samples at test time. The core principle leverages entropy minimization and pseudo-labeling to adapt the pre-trained model to the target participant's specific physiological patterns. We employ an adaptive selection mechanism that determines whether to apply TTA based on prediction confidence and distribution shift detection, ensuring robust performance across diverse participant populations.

\subsubsection{Adaptive Selection Mechanism}
We propose an adaptive Test-Time Adaptation (TTA) framework that intelligently decides whether to apply TTA for each test participant, as indiscriminate TTA application can degrade performance. Selective Test-Time Adaptation offers three advantages over universal adaptation. First, it prevents performance degradation for well-aligned participants where additional adaptation causes overfitting. Second, it reduces computational cost by adapting only high domain-shift participants, achieving latency reduction in our experiments. Third, it avoids error accumulation from unreliable pseudo-labels on well-matched participants, improving overall robustness. Our mechanism employs a hierarchical decision cascade with four stages. For participants with sufficient historical data ($n \geq 3$ experiments), we compute the average performance change and apply TTA if the average improvement exceeds $2\%$, or explicitly avoid it if performance degraded by more than $5\%$ in previous runs. Third, for participants without sufficient history, we quantify distribution shift between training and test data using three complementary metrics: Maximum Mean Discrepancy (MMD) measuring differences in feature means, KL Divergence quantifying distribution dissimilarity, and Variance Ratio capturing relative variability. These metrics are combined into a weighted distribution shift score $S_{\text{dist}}$, where TTA is skipped if $S_{\text{dist}} < 0.3$ indicating minimal distribution shift. Fourth, when distribution shift is substantial ($S_{\text{dist}} > 0.6$) or historical data is scarce, we conduct a quick performance test by applying three epochs of entropy minimization TTA on a validation subset and measuring the improvement rate; TTA is applied only if improvement exceeds $2\%$. 

\subsection{Sliding Window}
Our overlapping sliding window method processes continuous wearable device data streams using two components: a history window containing historical data for model input, and an output window representing the future prediction period. In this work, the input windows contain hhistotical data from Garmin watch data, in which $X = \{x_1, x_2, ..., x_T\}$ are the variables such as heart rate, activity levels, sleep conditions, where each $x_t \in \mathbb{R}^{15}$ represents the 15 selected features from the Garmin watch at time step $t$, and $T$ is the total number of time steps. We define $w_{in}$ as the input window length and $w_{out}$ as the prediction window length. The output window would then represent the stress levels to be predicted. For each time step $t$, we have: $X_{in}^t = \{x_{t-w_{in}+1}, x_{t-w_{in}+2}, ..., x_t\}$ and $Y_{out}^t = \{y_{t+1}, y_{t+2}, ..., y_{t+w_{out}}\}$, where $X_{in}^t$ is the input window at time $t$, and $Y_{out}^t$ is the corresponding output window to be predicted. The forecasting task can then be formulated as learning a function $f$ such that: $\hat{Y}_{out}^t = f(X_{in}^t)$, where $\hat{Y}_{out}^t$ is the predicted stress levels for the output window. Combinations of different input training windows (3, 5, 7, 9 days) and forecasting windows (1, 3, 5, 7 days) were tested in this study to find out the optimal training and forecasting window sizes. The sliding window algorithm starts with the first window of data, and then moves the window forward by a fixed step size with a step of one day. This creates overlapping consecutive windows, offering key advantages: capturing temporal dependencies, generating more training samples, and enabling continuous prediction.


\subsection{Leave-One-Out Cross-Validation (LOO-CV)}

In this study, we employ the Leave-One-Out Cross-Validation (LOO-CV) method to assess the generalization ability of our stress forecasting model. This approach is tailored to our dataset, which contains time series data from 16 participants. We $S = \{s_1, s_2, ..., s_{n}\}$ denote the set of n participants. The data for each participant $s$ is represented as $D_s = \{(x_1^s, y_1^s), (x_2^s, y_2^s), ..., (x_{N_s}^s, y_{N_s}^s)\}$, where $N_s$ is the number of samples for participant $s$, $x_i^s$ represents the input features, and $y_i^s$ the corresponding stress level. For each iteration of our cross-validation procedure, we partition the participants into three distinct sets: $s_{test} \in S$ (Test participant), $s_{val} \in S \setminus \{s_{test}\}$ (Validation participant), and $S_{train} = S \setminus \{s_{test}, s_{val}\}$ (Training participants). We then define our dataset partitions as follows: $D_{test} = D_{s_{test}}$, $D_{val} = D_{s_{val}}$, and $D_{train} = \bigcup_{s \in S_{train}} D_s$. The model is trained on $D_{train}$, tuned using $D_{val}$, and evaluated on $D_{test}$. The overall performance is computed as the average across all n iterations:

\begin{equation}
    \text{Performance} = \frac{1}{n} \sum_{i=1}^{n} \text{Performance}(D_{test}^i)
\end{equation}

where $D_{test}^i$ represents the test set for the $i$-th iteration.

In our implementation, we conduct 16 iterations of cross-validation, corresponding to our study's 16 participants. For each iteration, we select one participant as the test participant and randomly choose another as the validation participant. The remaining 14 participants' data are used for training. This process ensures that each participant serves as the test participant exactly once, providing a comprehensive evaluation of our model's performance across all individuals in our study. This approach simulates a real-world scenario where the model would be applied to new, unseen individuals, enhancing the practical relevance of our results. 


\subsection{Evaluation metrics}
\noindent
Six different metrics are used to evaluate performance and explainability. We use Mean Squared Error (MSE), Mean Absolute Error (MAE),  and Root Mean Squared Error (RMSE) as metrics of performance in this work because they are frequently adopted in forecasting tasks, particularly in deep learning contexts, due to their interpretability and computational simplicity. They are defined as:
\begin{equation}
\text{MSE} = \frac{1}{n} \sum_{i=1}^n (y_i - \hat{y}_i)^2,
\end{equation}
\begin{equation}
\text{MAE} = \frac{1}{n} \sum_{i=1}^n |y_i - \hat{y}_i|,
\end{equation}
\begin{equation}
\text{RMSE} = \sqrt{\frac{1}{n} \sum_{i=1}^n (y_i - \hat{y}_i)^2},
\end{equation}

where $y_i$ denotes the normalized stress value on day $i$, $\hat{y}_i$ represents the corresponding predicted normalized value, $\bar{y}$ is the mean of the actual values, and $n$ is the total number of days.

To assess the overall agreement between predicted and actual stress levels, we computed the Pearson correlation coefficient between the predicted values $\hat{\mathbf{y}} = (\hat{y}_1, \hat{y}_2, \ldots, \hat{y}_N)$ and true values $\mathbf{y} = (y_1, y_2, \ldots, y_N)$:
\begin{equation}
r = \frac{\sum_{i=1}^{N}(\hat{y}_i - \bar{\hat{y}})(y_i - \bar{y})}{\sqrt{\sum_{i=1}^{N}(\hat{y}_i - \bar{\hat{y}})^2}\sqrt{\sum_{i=1}^{N}(y_i - \bar{y})^2}}
\end{equation}
where $\bar{\hat{y}}$ and $y_i$ represent the mean of predicted and true stress levels, respectively. A high correlation coefficient indicates that the model captures the overall patterns and fluctuations in stress levels, which is essential for understanding individual stress dynamics and personalizing interventions.

The metric of the trend direction accuracy evaluates whether the model correctly predicts the direction of stress changes between consecutive time points. For each pair of consecutive measurements at time $t$ and $t+1$, we compare the signs of predicted and actual stress changes:
\begin{equation}
\text{TDA} = \frac{1}{N-1} \sum_{t=1}^{N-1} \mathbb{1}[\text{sign}(\hat{y}_{t+1} - \hat{y}_t) = \text{sign}(y_{t+1} - y_t)]
\end{equation}
where $\hat{y}_t$ and $y_t$ denote the predicted and true stress levels, and $\mathbb{1}[\cdot]$ is the indicator function. 


We also introduced the inverse coefficient of variation metric, $1/\text{CV} = \mu/\sigma$, where $\mu$ and $\sigma$ denote the mean and standard deviation of feature importance across participants. Higher values indicate greater stability across individuals. 

\begin{algorithm}[H]
\caption{our Model with Selective Domain Adaptation}
\label{alg:our-model}
  \footnotesize
  \textbf{Input:} $\mathcal{D}_{\text{train}}, \mathcal{D}_{\text{val}},
  \mathcal{D}_{\text{test}}$, Epochs $E=350$, Domain weight $\alpha \in
  [0,1]$
  \textbf{Output:} Trained model $\mathcal{M}$
  \begin{algorithmic}[1]
  \STATE \textbf{Phase 1: Training with Domain Adaptation}
  \STATE Initialize $\mathcal{M}$ (AdaptStress: Feature-level +
  Temporal Self-Attention + Domain Classifier)
  \STATE Initialize Adam optimizer with $\text{lr} = 5 \times 10^{-4}$,
  cosine warmup scheduler
  \FOR{epoch $= 1$ to $E$}
    \FOR{batch $(X, y, d) \in \mathcal{D}_{\text{train}}$}
        \STATE $\hat{y}, \hat{d} \leftarrow \mathcal{M}(X,
  \text{return\_domain}=\text{True})$
        \STATE $\mathcal{L}_{\text{main}} \leftarrow \text{MSE}(\hat{y},
  y)$
        \STATE $\mathcal{L}_{\text{domain}} \leftarrow
  \text{CrossEntropy}(\hat{d}, d)$
        \STATE $\mathcal{L} \leftarrow \mathcal{L}_{\text{main}} + \alpha
   \cdot \mathcal{L}_{\text{domain}}$
        \STATE Update $\mathcal{M}$ via backpropagation
    \ENDFOR
    \STATE Evaluate on $\mathcal{D}_{\text{val}}$ and apply Early
  Stopping (patience=30)
    \STATE Apply cosine learning rate scheduling
  \ENDFOR
  \STATE \textbf{Phase 2: Adaptive Test-Time Adaptation}
  \STATE Initialize TTA optimizer with $\text{lr}_{\text{tta}} = 10^{-4}$
  \STATE Compute domain shift metrics and adaptation threshold
  \IF{domain\_shift $>$ 0.3 AND performance\_improvement\_potential $>$ 2\%}
    \FOR{epoch $= 1$ to $E_{\text{tta}}=10$}
      \FOR{batch $(X_{\text{test}}, -, -) \in \mathcal{D}_{\text{test}}$}
          \STATE Generate $X_{\text{aug1}}, X_{\text{aug2}}$ with
  Gaussian noise ($\sigma_1=0.01, \sigma_2=0.02$)
          \STATE $\hat{y}_{\text{orig}} \leftarrow
  \mathcal{M}(X_{\text{test}})$
          \STATE $\hat{y}_{\text{aug1}} \leftarrow
  \mathcal{M}(X_{\text{aug1}})$
          \STATE $\hat{y}_{\text{aug2}} \leftarrow
  \mathcal{M}(X_{\text{aug2}})$
          \STATE $\mathcal{L}_{\text{consistency}} \leftarrow
  \frac{1}{3}\sum_{i,j} \text{MSE}(\hat{y}_i, \hat{y}_j)$ for all pairs
          \STATE Update $\mathcal{M}$ via backpropagation with
  $\mathcal{L}_{\text{consistency}}$
      \ENDFOR
    \ENDFOR
  \ELSE
    \STATE Skip TTA (insufficient domain shift or improvement potential)
  \ENDIF
  \STATE \textbf{Phase 3: Final Inference}
  \STATE $\mathcal{M}.\text{eval}()$
  \FOR{batch $(X_{\text{test}}, y_{\text{test}}, -) \in
  \mathcal{D}_{\text{test}}$}
    \STATE $\hat{y}_{\text{final}} \leftarrow
  \mathcal{M}(X_{\text{test}})$
  \ENDFOR
  \STATE Compute RMSE, MAE, MSE, Normalized MASE
  \end{algorithmic}
\end{algorithm}

\subsection{Explainability Analysis}
We implement SHAP (SHapley Additive exPlanations) analysis~\cite{lundberg2017unified} using KernelExplainer to interpret our deep learning models. To handle three-dimensional input data (samples × time steps × features), we apply temporal aggregation with 50 background samples and 100 test samples for computational efficiency.
Our SHAP implementation provides feature importance rankings and reveals how each factor influences stress quality predictions. 


\begin{figure*}[!t]
    \centering
    \includegraphics[width=1\textwidth]{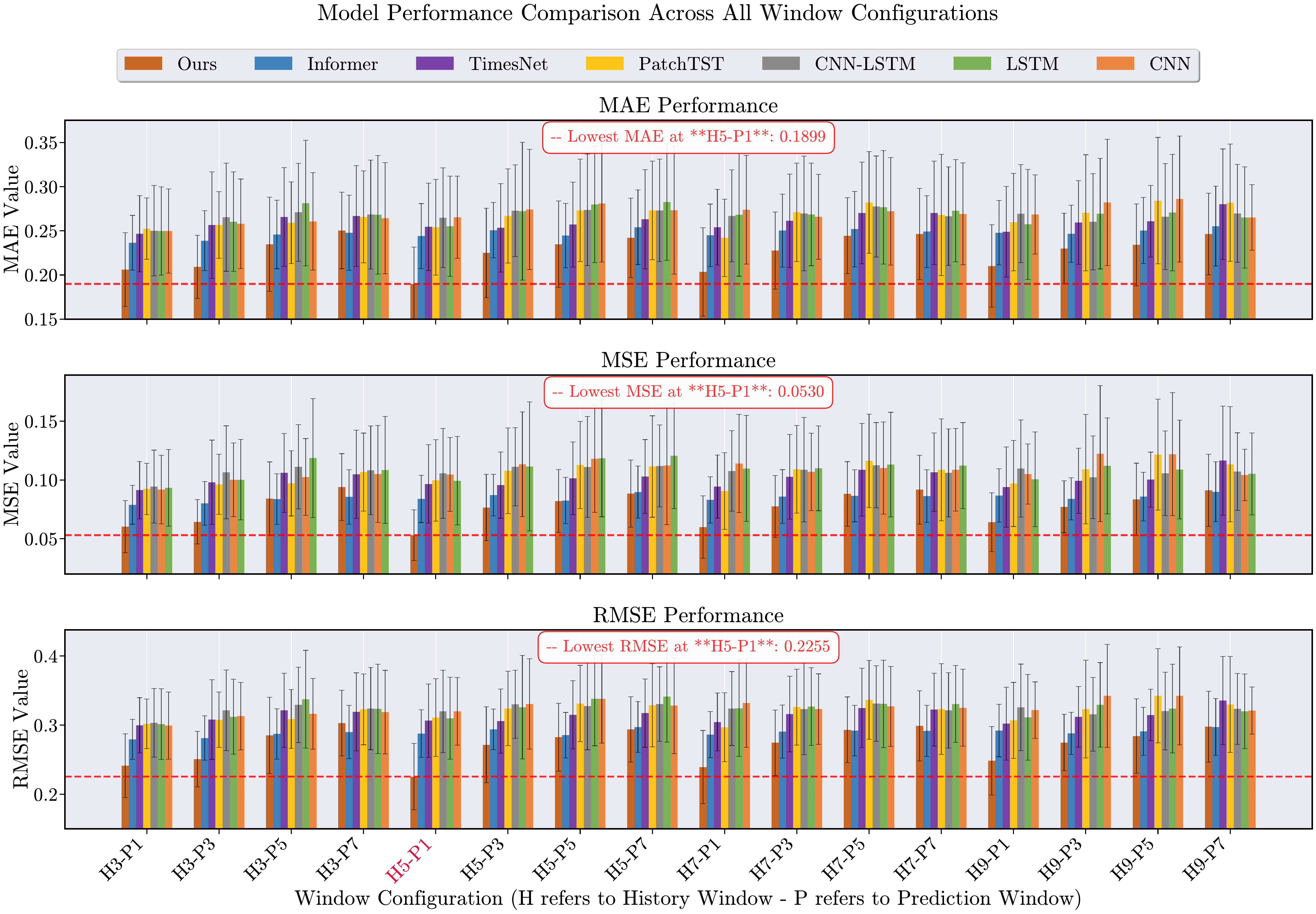}
    \caption{Comprehensive performance comparison across all history window sizes (3, 5, 7, 9 days) and prediction horizons (1, 3, 5, 7 days) for seven different forecasting models: our proposed approach, state-of-the-art time series models (Informer, TimesNet, PatchTST), and traditional deep learning baselines (CNN-LSTM, LSTM, CNN) by three evaluation metrics (Mean Absolute Error, Mean Squared Error, Root Mean Squared Error).}
    \label{fig:allwindows}
\end{figure*}

\subsection{Computational Infrastructure}
\label{sec:infrastructure}
Our experimental framework utilized a dual-system configuration to balance development efficiency and computational demands. Initial model prototyping and validation were performed on a dedicated research server equipped with an Intel i7-13700K processor, 128 GB memory, and an NVIDIA RTX 4090 graphics card. For computationally intensive tasks, particularly exhaustive hyperparameter optimization, we leveraged institutional high-performance computing resources through the Hábrók Cluster\footnote{\url{https://wiki.hpc.rug.nl/habrok/introduction/cluster_description}}. This infrastructure provides access to 190 computing nodes with diverse architectures: 119 AMD EPYC-based standard nodes (128 cores each), 4 Intel Xeon high-memory systems (80 cores, 4 TB memory), and specialized GPU-accelerated nodes incorporating NVIDIA's A100, V100, L40s, and H100 processors. The entire experimental pipeline was developed using the PyTorch framework to ensure reproducibility and portability across different computing environments.

\begin{table*}[htbp]
\centering
\caption{Comprehensive performance comparison across different metrics, input, models, and prediction window sizes}
\label{tab:model_comparison_all_metrics}
\resizebox{1.01\textwidth}{!}{
\scriptsize 
\begin{tabular}{l|l|c|c|c|c||c|c|c|c||c|c|c|c||c|c|c|c}
\toprule[1.5pt]
\multirow{2}{*}{\textbf{Metric}} & \multirow{2}{*}{\textbf{Model}} & \multicolumn{4}{c||}{\textbf{Input = 3 days}} & \multicolumn{4}{c||}{\textbf{Input = 5 days}} & \multicolumn{4}{c||}{\textbf{Input = 7 days}} & \multicolumn{4}{c}{\textbf{Input = 9 days}} \\
\cline{3-18}
 & & 1 & 3 & 5 & 7 & 1 & 3 & 5 & 7 & 1 & 3 & 5 & 7 & 1 & 3 & 5 & 7 \\
 
\midrule
\multirow{7}{*}{\textbf{MAE}} 

& LSTM & 0.250 & 0.260 & 0.281 & 0.268 & 0.255 & 0.272 & 0.280 & 0.283 & 0.268 & 0.269 & 0.279 & 0.273 & 0.257 & 0.269 & 0.271 & 0.265 \\
 & CNN & 0.238 & 0.247 & 0.256 & 0.266 & 0.244 & 0.252 & 0.263 & 0.274 & 0.254 & 0.260 & 0.261 & 0.268 & 0.253 & 0.263 & 0.261 & 0.271 \\
 & CNN-LSTM & 0.238 & 0.260 & 0.255 & 0.261 & 0.247 & 0.250 & 0.254 & 0.268 & 0.257 & 0.256 & 0.261 & 0.260 & 0.251 & 0.264 & 0.254 & 0.266 \\
 & Informer & 0.237 & 0.239 & 0.246 & 0.248 & 0.244 & 0.251 & 0.245 & 0.254 & 0.245 & 0.250 & 0.252 & 0.249 & 0.248 & 0.247 & 0.250 & 0.255 \\
 & PatchTST & 0.252 & 0.257 & 0.259 & 0.266 & 0.254 & 0.267 & 0.273 & 0.273 & 0.261 & 0.271 & 0.282 & 0.268 & 0.254 & 0.270 & 0.284 & 0.282 \\
 & TimesNet & 0.247 & 0.257 & 0.266 & 0.267 & 0.254 & 0.253 & 0.257 & 0.263 & 0.254 & 0.261 & 0.270 & 0.270 & 0.249 & 0.239 & 0.281 & 0.280 \\
& \textbf{Ours} & 0.197 & 0.266 & 0.266 & 0.308 & \textbf{0.190} & 0.371 & 0.386 & 0.277 & 0.204 & 0.362 & 0.294 & 0.278 & 0.197 & 0.370 & 0.308 & 0.395 \\
 
\midrule
\multirow{7}{*}{\textbf{MSE}} 
 & LSTM & 0.093 & 0.100 & 0.119 & 0.109 & 0.099 & 0.111 & 0.119 & 0.121 & 0.110 & 0.110 & 0.113 & 0.112 & 0.101 & 0.112 & 0.109 & 0.105 \\
 & CNN & 0.085 & 0.090 & 0.098 & 0.104 & 0.089 & 0.095 & 0.103 & 0.110 & 0.097 & 0.100 & 0.102 & 0.103 & 0.097 & 0.102 & 0.100 & 0.108 \\
 & CNN-LSTM & 0.085 & 0.101 & 0.098 & 0.100 & 0.091 & 0.090 & 0.094 & 0.105 & 0.098 & 0.094 & 0.101 & 0.095 & 0.093 & 0.104 & 0.093 & 0.104 \\
 & Informer & 0.079 & 0.080 & 0.084 & 0.086 & 0.084 & 0.087 & 0.082 & 0.090 & 0.083 & 0.086 & 0.087 & 0.086 & 0.087 & 0.084 & 0.086 & 0.090 \\
 & PatchTST & 0.092 & 0.096 & 0.097 & 0.107 & 0.100 & 0.108 & 0.113 & 0.111 & 0.091 & 0.109 & 0.116 & 0.109 & 0.097 & 0.109 & 0.122 & 0.113 \\
 & TimesNet & 0.091 & 0.098 & 0.106 & 0.105 & 0.097 & 0.096 & 0.101 & 0.103 & 0.094 & 0.103 & 0.109 & 0.107 & 0.096 & 0.099 & 0.100 & 0.116 \\
& \textbf{Ours} & 0.057 & 0.109 & 0.110 & 0.150 & \textbf{0.053} & 0.105 & 0.124 & 0.117 & 0.061 & 0.201 & 0.130 & 0.116 & 0.058 & 0.207 & 0.145 & 0.237 \\

\midrule
\multirow{7}{*}{\textbf{RMSE}} 
 & LSTM & 0.302 & 0.312 & 0.338 & 0.323 & 0.312 & 0.310 & 0.326 & 0.338 & 0.341 & 0.324 & 0.327 & 0.331 & 0.331 & 0.330 & 0.324 & 0.320 \\
 & CNN & 0.299 & 0.313 & 0.316 & 0.319 & 0.320 & 0.330 & 0.338 & 0.328 & 0.331 & 0.323 & 0.327 & 0.325 & 0.322 & 0.342 & 0.332 & 0.321 \\
 & CNN-LSTM & 0.303 & 0.321 & 0.329 & 0.324 & 0.320 & 0.330 & 0.328 & 0.330 & 0.323 & 0.323 & 0.331 & 0.321 & 0.326 & 0.316 & 0.320 & 0.323 \\
 & Informer & 0.279 & 0.281 & 0.287 & 0.290 & 0.288 & 0.294 & 0.285 & 0.297 & 0.286 & 0.291 & 0.292 & 0.292 & 0.292 & 0.288 & 0.291 & 0.297 \\
 & PatchTST & 0.302 & 0.308 & 0.309 & 0.323 & 0.311 & 0.324 & 0.332 & 0.329 & 0.297 & 0.326 & 0.337 & 0.323 & 0.307 & 0.323 & 0.342 & 0.330 \\
 & TimesNet & 0.300 & 0.308 & 0.321 & 0.319 & 0.306 & 0.306 & 0.315 & 0.317 & 0.305 & 0.316 & 0.325 & 0.322 & 0.302 & 0.312 & 0.315 & 0.334 \\
 & \textbf{Ours} & 0.241 & 0.251 & 0.285 & 0.303 & \textbf{0.226} & 0.272 & 0.283 & 0.294 & 0.239 & 0.275 & 0.293 & 0.298 & 0.248 & 0.275 & 0.284 & 0.298 \\
\bottomrule[1.5pt]
\end{tabular}
}
\end{table*}

\section{Experimental Results and Analysis}
Figure~\ref{fig:allwindows} presents a comprehensive comparison of all models' performance across 16 different window configurations, combining four history window sizes (H3, H5, H7, H9) with four prediction horizons (P1, P3, P5, P7). Our proposed method consistently outperforms all baseline models across all experimental configurations. The optimal performance is achieved at the H5-P1 configuration, with MAE of 0.1899, MSE of 0.0530, and RMSE of 0.2255, demonstrating the model's capability to capture stress dynamics effectively with a moderate history window and short-term prediction horizon. Among the baseline models, Informer demonstrates the most competitive performance, followed by TimesNet, PatchTST, CNN-LSTM, LSTM, and CNN.
Two important trends emerge from the comparative analysis. First, all models exhibit consistent degradation in prediction accuracy as the forecasting horizon extends from P1 to P7, highlighting the inherent challenge of long-term stress forecasting from wearable sensor data. This performance deterioration is more pronounced in baseline models compared to our approach, indicating superior robustness of our method for extended predictions. Second, the history window size (H3 to H9) exhibits a relatively modest impact on model performance compared to the prediction horizon, suggesting that stress patterns can be effectively captured within shorter temporal contexts when using appropriate modeling techniques. The error bars in the figure further reveal that our proposed method shows notably smaller variance across the 16 participants compared to baseline models, demonstrating superior generalization capability and robustness across individuals with diverse stress patterns and physiological responses.

Table~\ref{tab:model_comparison_all_metrics} presents comprehensive experimental results comparing our proposed model against six baseline methods across varying input windows (3-9 days) and prediction horizons (1-7 days). Our model achieves superior performance in most configurations. While performance naturally degrades with longer prediction horizons—MSE increasing from 0.053 to 0.105 when extending from 1-day to 5-day predictions—our approach maintains consistent advantages across different metrics. Among baselines, Informer demonstrates the most competitive results (MSE: 0.079-0.090), outperforming both traditional models (LSTM, CNN) and more recent transformer-based architectures (PatchTST, TimesNet), suggesting that model complexity alone does not guarantee improved performance for stress forecasting tasks. The results validate that a 5-day input window provides optimal temporal context for capturing stress patterns.

\begin{figure*}[t]
    \centering
    \includegraphics[width=1\textwidth]{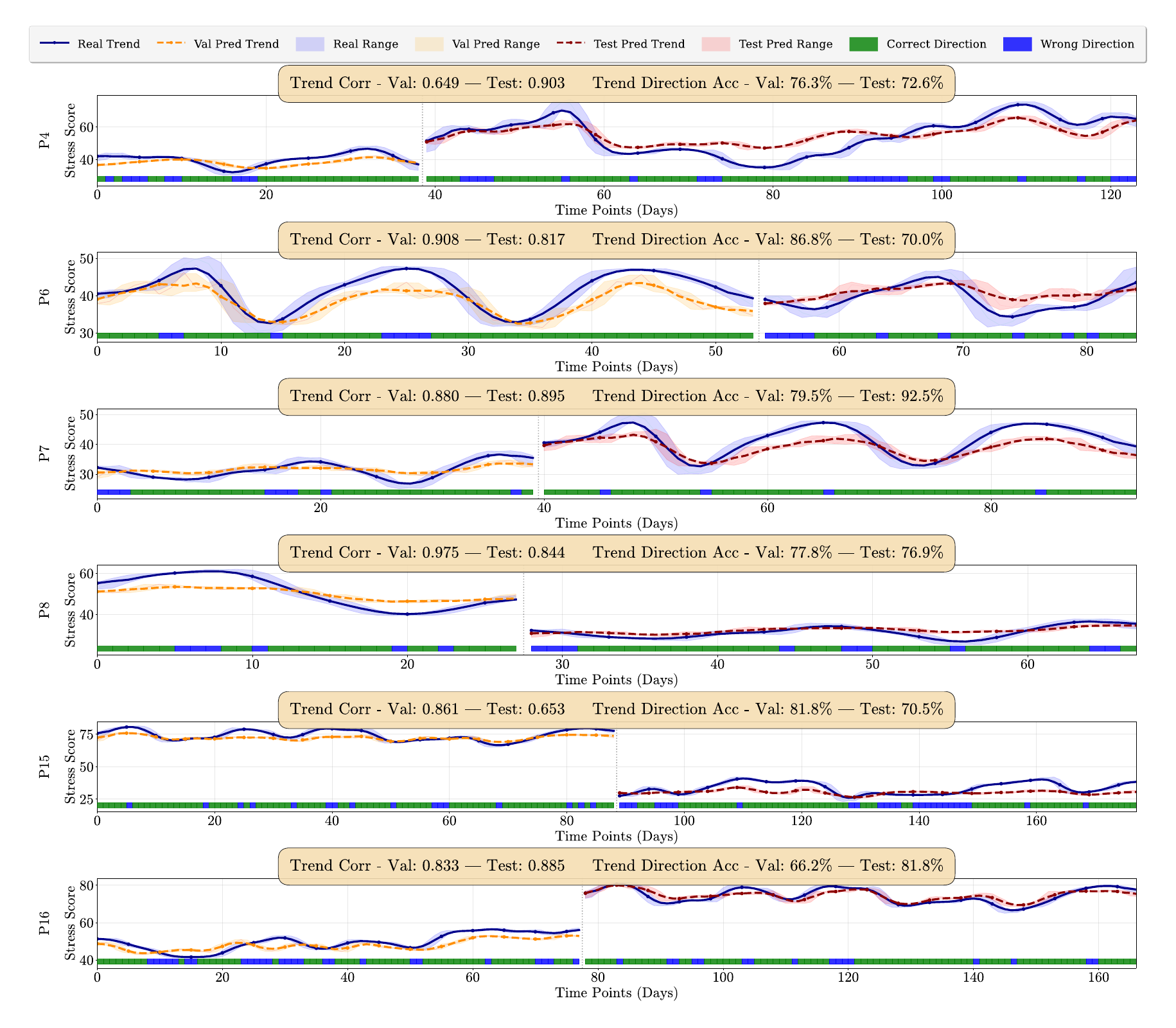}
    \caption{Comparison of ground truth and predicted stress values for eight representative participants. Solid blue lines indicate actual stress scores, with yellow dotted lines showing validation predictions and red lines showing test predictions. Shaded regions represent prediction confidence intervals. Bottom bars indicate trend direction accuracy (green: correct, blue: incorrect).}
    \label{fig:real-predic}
\end{figure*}

\subsection{Visualization of representative Participant: Prediction vs True value of stress}
Figure~\ref{fig:real-predic} presents comprehensive results from our stress prediction experiments across eight representative participants (P4, P6, P7, P8, P15, P16). These participants exemplify diverse stress profiles, including high versus low baseline stress levels, stable versus volatile patterns, gradual versus sudden stress transitions, and varying monitoring durations, thereby providing a comprehensive yet concise representation of the model's performance across the entire cohort. Our results demonstrate the model's ability to predict stress scores using physiological data collected from Garmin smartwatches. The visualization employs multiple visual elements to convey model performance: solid blue lines represent ground truth stress scores, dotted yellow lines show validation predictions, red lines indicate test set predictions, with corresponding shaded regions illustrating prediction confidence intervals. Additionally, the bottom bars encode trend direction indicators, where green represents correct direction predictions and red highlights incorrect predictions.
\begin{figure}[!t]
    \centering
    \includegraphics[width=0.46\textwidth]{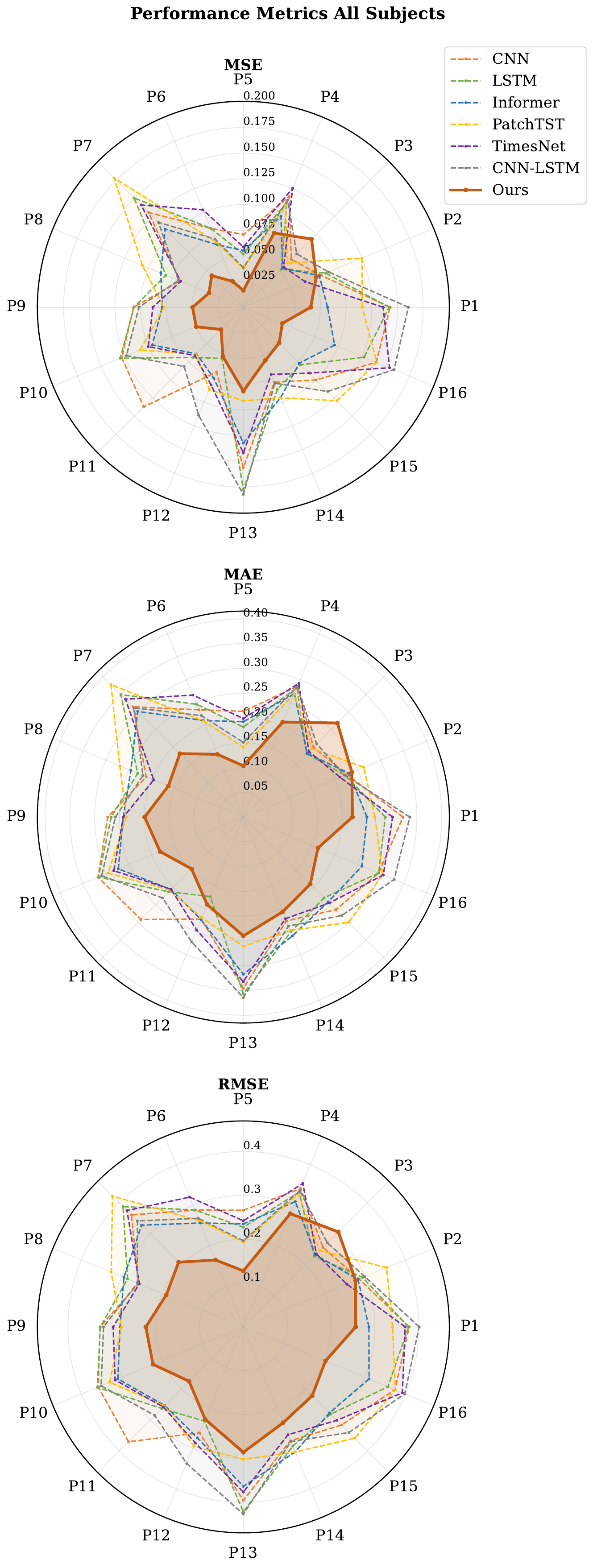}
    \caption{Radar plot comparison of prediction errors (MAE, MSE, RMSE) across 16 participants for all seven models (H5-P1). Smaller areas indicate better performance.}
    \label{fig:metric_comparison}
\end{figure}
Trend correlation coefficients demonstrate strong positive relationships, with validation correlations ranging from 0.649 to 0.975 (mean: 0.851) and test correlations spanning 0.653 to 0.903 (mean: 0.827). Particularly noteworthy is the trend direction accuracy, which achieves validation accuracies between 66.2\% and 86.8\%, with test accuracies reaching an impressive 92.5\% for participant P7. 


\begin{figure*}
    \centering
    \includegraphics[width=1\textwidth]{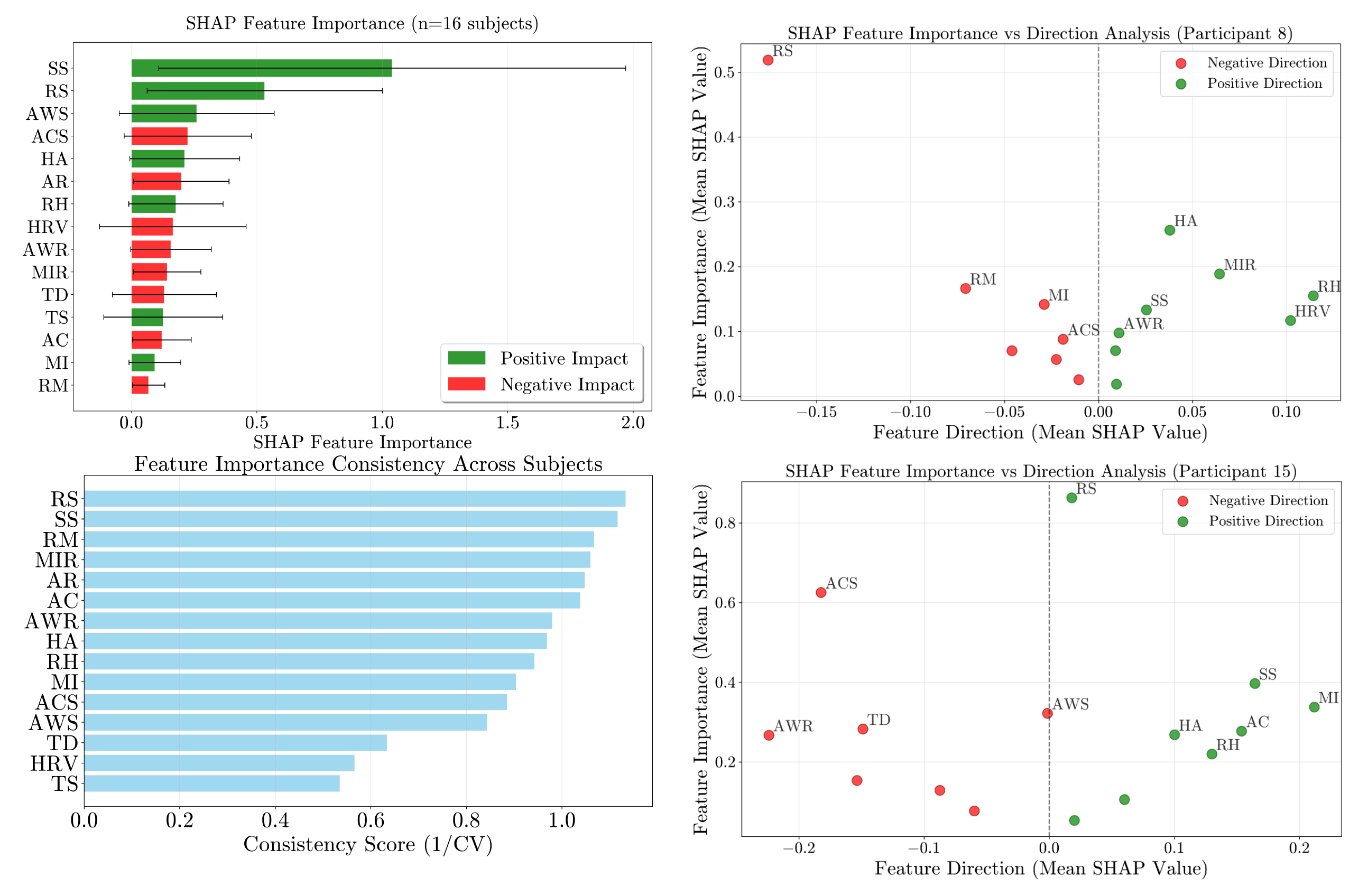}
    \caption{Feature importance, consistency, and individual-specific patterns in stress prediction revealed through SHAP analysis.}
    \label{fig:feature-importance}
\end{figure*}

As shown in Figure~\ref{fig:real-predic}, comparison between true and predicted stress values from different participants indicates that our model captures both short-term fluctuations and long-term stress patterns. Participants P4, P7, and P16, monitored over approximately 80 days, exhibit complex stress profiles with multiple peaks and valleys that the model tracks with high fidelity. The model demonstrates particular strength in capturing gradual stress buildups (observable in P7 around day 60-80) and sudden stress releases (evident in P4 around day 50). The ability to track these diverse temporal patterns suggests that the model has learned meaningful patterns that generalize across different stress manifestation patterns. 

The prediction uncertainty bands provide valuable insights into model confidence and reliability. Narrower bands during stable stress periods (e.g., P8 days 30-50) indicate high model confidence, while wider bands during volatile periods (e.g., P4 days 50-60) appropriately reflect increased uncertainty. This adaptive uncertainty quantification is crucial for clinical applications, allowing healthcare providers to gauge prediction reliability. The model demonstrates a sophisticated understanding of its own limitations—participants with more stable stress patterns (P8, P15) show consistently narrow uncertainty bands, while those with higher variability (P4, P6, and P7) exhibit appropriately wider confidence intervals. 


Individual participant analysis reveals patterns that highlight both the model's adaptability and the heterogeneity in stress responses. Participant P7 achieves the highest test correlation (0.895) with 92.5\% trend direction accuracy. Conversely, P15 presents the most challenging case with the lowest validation correlation (0.652), yet remarkably achieves 70.5\% test trend direction accuracy. This apparent paradox suggests that even when absolute value predictions are less accurate, the model maintains a strong capability in capturing stress dynamics and directional changes.

The model's consistent performance across diverse monitoring durations deserves special attention. Participants with shorter monitoring periods (P6 and P15, approximately 90 days) achieve comparable performance to those with extended monitoring (P8, over 180 days), with correlation coefficients remaining stable throughout the observation periods. This temporal stability indicates that the model does not suffer from performance degradation over time. Furthermore, the model adapts well to different baseline stress levels—from participants with low baseline stress around 25-30 (P15) to those with elevated baselines near 50-70 (P4, P7)—maintaining consistent relative accuracy across this range.

The trend direction accuracy metrics provide evidence for the model's practical utility. With test accuracies ranging from 70.0\% to 92.5\% (mean: 77.4\%), the model demonstrates reliable capability in predicting whether stress will increase or decrease. This is evidenced by the predominance of green indicators in the direction bars across all participants, with extended green segments during periods of consistent trends. The model's ability to correctly identify trend changes is crucial for preventive stress management, as it enables timely interventions before stress levels reach critical thresholds. Notably, even in challenging cases like P15, where absolute correlation is moderate, the high trend direction accuracy (70.5\%) ensures practical utility for stress trajectory monitoring.


\subsection{Participant-Specific Performance Analysis}

To evaluate the generalizability and robustness of our proposed method across different individuals, we conducted a comprehensive participant-specific analysis using data collected from Garmin wearable devices. Figure~\ref{fig:metric_comparison} presents radar charts displaying the performance metrics (MAE, MSE, and RMSE) for all seven models across 16 participants (P1-P16) under the optimal window configuration (History window size 5 days, Prediction window size 1 day).

Our proposed method demonstrates superior performance across most participants, as evidenced by the orange areas residing closest to the center of each radar chart, indicating the lowest error values. The model achieves stable performance with MSE values ranging from approximately 0.025 to 0.075, MAE values between 0.1 and 0.20, and RMSE values spanning 0.10 to 0.25 across all participants.

Notably, significant inter-participant variability is observed in prediction difficulty, with certain participants (e.g., P5, P11, P16) showing consistently lower error rates across all models, while others (e.g., P4, P9, P13) present more challenging prediction scenarios. 


Among the baseline methods, Informer (blue dashed line) generally exhibits the second-best performance, followed by TimesNet (purple dashed line), PatchTST (yellow dashed line), and LSTM (green dashed line). However, the performance gaps between baseline models vary considerably across participants, suggesting that individual physiological characteristics may favor different modeling approaches. Despite this variability, our proposed method maintains consistent superiority across all participants, demonstrating its robust capability to adapt to diverse individual patterns in wearable-based stress prediction.


\subsection{Model Interpretability Analysis}

To understand the decision-making process of our stress prediction model, we conducted a comprehensive explainability analysis using SHAP (SHapley Additive exPlanations) across multiple levels: population-averaged patterns, cross-participant consistency, and individual-specific variations. Figure~\ref{fig:feature-importance} presents these complementary perspectives, revealing how various physiological features contribute to stress predictions.

\subsubsection{Population-Level Feature Importance}

The top-left panel of Figure \ref{fig:feature-importance} displays averaged feature importance across all 16 participants, revealing distinct patterns in how physiological markers influence stress predictions. Sleep-related features emerge as dominant predictors, with Sleep Overall Score (SS) achieving the highest importance value (1.1) and REM Sleep Seconds (RS) value(0.5). The prominence of sleep metrics as primary stress indicators aligns with established research linking sleep quality to psychological well-being.

Interestingly, the analysis reveals bidirectional feature impacts that reflect the complex nature of stress physiology. Features showing negative impacts—including Active Seconds (ACS), Average Respiration (AR), Highest Respiration Value (HRV), Average Walking Respiration Value (AWR), Minimum Average Heart Rate (MIR), and Total Distance Meters (TD)—suggest protective effects where higher values correlate with reduced stress predictions. Meanwhile, features such as Awake Sleep Seconds (AWS) and Resting Heart Rate (RH) demonstrate positive but moderate contributions with smaller error bars, suggesting consistent but subtle roles in stress prediction.

\subsubsection{Cross-participant Consistency Analysis}

The bottom-left panel quantifies feature consistency across participants using the inverse coefficient of variation (1/CV), where higher values indicate greater stability. This analysis reveals a clear hierarchy in feature reliability: sleep metrics (RS and SS) demonstrate exceptional consistency (0.9-1.0), establishing them as universal stress indicators regardless of individual differences. Conversely, activity-based features such as Total Steps (TS), Highest Respiration Value (HRV), and Total Distance Meters (TD) exhibit low consistency scores (0.1-0.2), highlighting substantial inter-participant variability.

Mid-range consistency scores (0.3-0.7) for features including AWS, ACS, MI, RH, and HA suggest moderate variability, while sleep disruption markers like Restless Moment Count (RM) maintain relatively high consistency (~0.8). This stratification indicates that our model successfully distinguishes between universal physiological patterns and individual-specific responses, enabling both population-level insights and personalized predictions.

\subsubsection{Individual-Specific Stress Patterns}

The right panels of Figure~\ref{fig:feature-importance} illustrate how identical physiological features can manifest opposite effects across individuals, exemplified through two representative participants. For Participant 8 (top-right), REM Sleep Seconds dominates with high importance (0.52) and strong negative direction (-0.15), confirming sleep as a primary stress reducer. Activity features (RM, MI, ACS) similarly cluster in the negative direction with moderate importance (0.05-0.17), while cardiovascular markers (MIR, HA, RH, HRV) show positive impacts (0.02-0.25).

Participant 15 (bottom-right) presents a strikingly different profile despite RS maintaining high importance (0.85). Critically, its directional impact reverses to positive (0.02), suggesting REM sleep correlates with increased stress for this individual—a complete inversion of the typical pattern. Additionally, Active Seconds (ACS) becomes the dominant stress reducer (-0.17, importance: 0.63), while features like SS paradoxically show positive impacts. These opposing patterns within the same physiological framework demonstrate that stress responses are highly individualized. 


\section{Conclusions}
Our experimental results demonstrate substantial improvements over state-of-the-art stress forecasting baselines, achieving optimal performance that represents improvements exceeding 20\% across all metrics compared to the best baseline model. The model maintains robust performance across all tested window configurations, with consistently high test correlations and trend direction accuracy for all participants. This consistent performance across extended monitoring periods without degradation validates the temporal stability essential for long-term deployment. Notably, the model's adaptive uncertainty quantification—narrowing confidence intervals during stable periods and appropriately widening during volatile episodes—provides the reliability necessary for clinical decision support.

The explainability analysis through SHAP reveals insights into feature relevance. Sleep metrics emerge as dominant and highly consistent predictors across participants, while activity features exhibit substantial inter-participant variability, indicating personalized stress responses. Most significantly, the model captures sophisticated individual-specific patterns where identical features can have opposing effects. 



The model's ability to capture both gradual stress buildups and sudden releases, while adapting to individual baseline variations (25-50 range), ensures applicability across diverse populations. Furthermore, the consistent performance across different monitoring durations and the robust generalization from validation to test sets (correlation improvement from 0.769 to 0.834 mean) confirm the model's readiness for real-world deployment.

This research establishes that consumer wearables, combined with interpretable machine learning, can democratize access to continuous stress predicting. Future work should expand to larger, more diverse populations and integrate contextual factors such as environmental and social variables. Furthermore, clinical studies which incoperate interventions that help users to cope with stress need to be conducted to investigate whether stress forecasting can assist in lifestyle interventions.

\vspace{0.2in}
\section*{Acknowledgments}
This work was supported by the Healthy Living as a Service project, funded by the Dutch Research Council (Nederlandse Organisatie voor Wetenschappelijk Onderzoek, NWO) under grant number 13948 as part of the KIC research programme. We gratefully acknowledge the Center for Information Technology at the University of Groningen for their support and for providing access to the Hábrók high-performance computing cluster.
\bibliographystyle{IEEEtran}

\bibliography{ref}









\end{document}